%% file: main.tex
\documentclass{article}

\PassOptionsToPackage{numbers}{natbib}

\usepackage[preprint]{neurips_data_2024}

\usepackage[numbers]{natbib}
\usepackage[utf8]{inputenc} %
\usepackage[T1]{fontenc}    %
\usepackage{hyperref}       %
\usepackage{url}            %
\usepackage{booktabs}       %
\usepackage{amsfonts}       %
\usepackage{nicefrac}       %
\usepackage{microtype}      %
\usepackage{xcolor}         %

\usepackage{inconsolata}

\usepackage{graphicx} 
\usepackage{multirow}
\usepackage{booktabs}
\usepackage{subfigure}
\usepackage{url}
\usepackage{float}
\usepackage{makecell}
\usepackage{tcolorbox}
\usepackage{arydshln}
\usepackage{adjustbox}
\usepackage{enumitem}
\usepackage[fixed]{fontawesome5}
\usepackage{pythonhighlight}
\usepackage{caption}
\usepackage{wrapfig}

\hypersetup{
    urlcolor=pink,
}

\newcommand{\DataName}{\textsc{SimulBench}}

\title{\DataName{}: Evaluating Language Models with Creative Simulation Tasks}

\author{Qi Jia, Xiang Yue, Tianyu Zheng, Jie Huang, Yuchen Lin}

\author{Qi Jia$^2$ \quad  Xiang Yue$^{3}$ \quad  Tianyu Zheng$^4$ \quad  Jie Huang$^5$ \quad  Bill Yuchen Lin$^1$\thanks{Corresponding author.}\\[5pt]
	$^1$Allen Institute for AI\quad $^2$National University of Singapore\quad $^3$Carnegie Mellon University\\ 
    $^4$University of Waterloo\quad
    $^5$University of Illinois, Urbana Champaign\\[5pt]
{\scriptsize{\faEnvelope[regular]}} \texttt{jia\_qi@nus.edu.sg \& yuchenl@allenai.org} \\[5pt]
 \quad {{\scriptsize \faStar[regular]} \texttt{\href{https://simulbench.github.io}{\textcolor{blue}{https://simulbench.github.io}}}}
}

\begin{document}

\maketitle

\begin{abstract}
We introduce \DataName{}, a benchmark designed to evaluate large language models (LLMs) across a diverse collection of creative simulation tasks, such as acting as a Linux terminal or playing text games with users. 
While these simulation tasks serve as effective measures of an LLM's general intelligence, they are seldom incorporated into existing benchmarks.
A major challenge is to develop an evaluation framework for testing different LLMs fairly while preserving the multi-round interactive nature of simulation tasks between users and AI. To tackle this issue, we suggest using a fixed LLM as a user agent to engage with an LLM to collect dialogues first under different tasks. Then, challenging dialogue scripts are extracted for evaluating different target LLMs. To facilitate automatic assessment on \DataName{}, GPT-4 is employed as the evaluator, tasked with reviewing the quality of the final response generated by the target LLMs given multi-turn dialogue scripts.
Our comprehensive experiments indicate that these creative simulation tasks continue to pose a significant challenge with their unique natures and show the gap between proprietary models and the most advanced open LLMs. For example, GPT-4-turbo outperforms LLaMA-3-70b-Chat on 18.55\% more cases.
\end{abstract}

\input{introduction}

\input{evaluation}

\input{results}

\input{conclusion}

\input{limitations}
\input{ethics}
\bibliographystyle{plain}
\bibliography{neurips_data_2024} 

\appendix
\input{appendix}

\end{document}

%% file: introduction.tex
\begin{figure}[ht!] 
    \centering
    \includegraphics[width=1.0\linewidth]{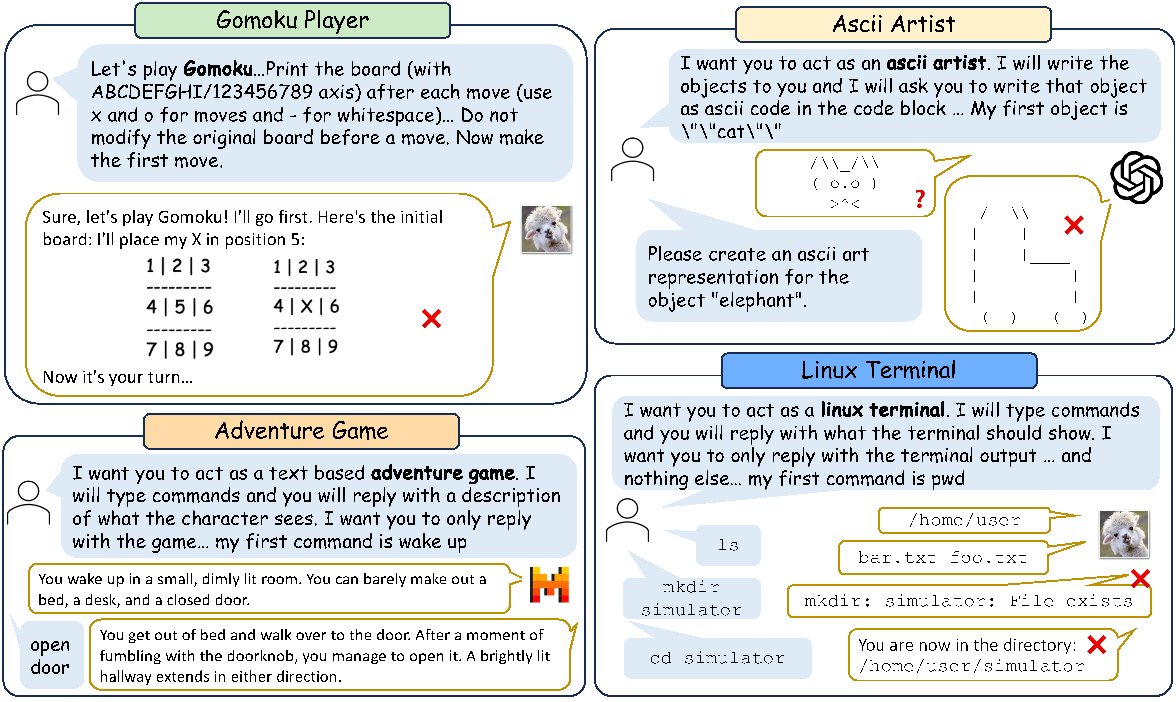} 
    \caption{Examples of creative simulation tasks in \DataName{}. \vspace{-1em}}
    \label{fig:examples} 
\end{figure} 

\section{Introduction}

The ability of large language models (LLMs) to simulate complex tasks is pivotal in driving the evolution of AI towards achieving general intelligence~\citep{bubeck2023sparks}. These models exhibit remarkable versatility by adeptly assuming a wide range of roles—from acting as a Linux terminal to serving as an investment manager—highlighting their adaptability across various domains. Such flexibility underscores their potential for broad implementation. Consequently, the development of a benchmark dataset for simulation tasks is imperative in nurturing LLMs' progression toward becoming true generalists.

Nonetheless, existing benchmarks do not fully evaluate this potential. Current evaluations mainly focus on single-turn, static interactions between users and LLMs~\citep{alpaca_eval,yue2023mmmu}. While MT-bench~\citep{zheng2023judging} attempts to consider multi-turn interactions with 80 examples, its reliance on \textit{predefined} second queries fails to effectively examine the dynamic responses of different LLMs when engaging with users in complex, long-horizon simulation tasks. In addition, these benchmarks primarily concentrate on tasks related to general information retrieval and creative writing, with less emphasis on complex simulation abilities.

Based on whether the simulation target is a human or not, simulation tasks can be divided into two groups. The former groups correlate to existing role-playing benchmarks focusing on replicating the language styles and knowledge of famous characters or professions~\citep{li2023chatharuhi,wang2023rolellm,zhou2023characterglm,shen2023roleeval} and have been widely investigated. However, the second group of tasks are under consideration. Recent work from Duan et al.~\cite{duan2024gtbench} introducing GTBench to explore LLM's ability on some language-driven games, is barely beginning to explore this kind of simulation abilities. A comprehensive benchmark covering wide-ranging non-human centered tasks for thoroughly assessing the simulation potential of LLMs is in urgent need.

\textbf{Tasks for \DataName{}.}
We have gathered 109 distinct simulation tasks that require LLMs to perform in a variety of interfaces. These interfaces include acting as a Linux terminal, an SQL executor, text-based games such as tic-tac-toe, a generator for passwords with particular constraints, an ASCII art creator, a predictor of chemical reactions, and more.  
Each task specification comes with an interface description, some output requirements and an initial user request. 
Some examples of simulation tasks are presented in Figure~\ref{fig:examples}.

\textbf{Multi-Turn Script-based Evaluation.} 
MT-bench~\citep{zheng2023judging} is designed to test LLMs in a two-turn conversation, where the second turn is predefined. However, our \DataName{} necessitates multiple turns between users and LLMs. Depending on the task types and context window limit, some tasks may involve conversations exceeding 5 turns, with the majority spanning over 2 turns. To replicate realistic usage scenarios of LLMs, we employ OpenAI's GPT-3.5 to simulate a user interacting continuously with an LLM. To ensure fairness among different test models, we extract challenging histories from the collected dialogues to form the final test scripts. Finally, after gathering reactions from each target LLM, we follow the methodology of previous studies ~\citep{zheng2023judging,zhou2023sotopia,zhang2023comprehensive}, using GPT-4 to assess and rate the quality of these responses. We also conduct pairwise comparisons for a more detailed analysis.

\textbf{Experimental Results and Findings.}
Our study involved an analysis of 2 proprietary LLMs and 12 widely used open-source LLMs, specifically series of models in LLaMA~\citep{touvron2023llama}, Qwen~\citep{bai2023qwen} and Mixtral%
~\citep{jiang2024mixtral}. These models are often ranked highly on several existing leaderboards, such as the Chatbot Arena\footnote{\url{https://huggingface.co/spaces/lmsys/chatbot-arena-leaderboard}}. Although the performance of these open-source LLMs is approaching GPT-4-turbo, there is still a conspicuous gap between them. Even the strongest open LLM, LLaMA-3-70B-Chat, was surpassed by GPT-4-turbo on 18.55\% more cases on the hard subset of \DataName{}.

We noticed that recent LLMs can take advantage of history information much better than the previous ones, showing superior performance on stateful tasks than the stateless ones. However, we also highlighted the importance of utilizing the context information cautiously and selectively, and showed that even the performance of GPT-4o drops from 9.40 to 7.57 on the most challenging scripts possibly containing erroneous dialogue history. In addition, we observed that although LLMs are knowledgeable and good at question answering, they face obstacles to applying knowledge flexibly and tend to exhibit poorer performance in simulation tasks that necessitate more rigorous outputs (such as classical encryption algorithms) and strategic plans along with long-horizon memory (such as different board games).

%% file: evaluation.tex
\section{\DataName{} }

In this section, we first describe how we collect the tasks for \DataName{}, next introduce the user agent for collecting LLM interactions, then explain how we extracted challenging conversation histories and finally present our evaluation metrics.

\subsection{Overview}
The complexity of simulation tasks, characterized by their multi-round nature and the diverse conversation paths influenced by differing model responses, renders script-based single-turn evaluation~\citep{zhao2023narrativeplay,wang2023rolellm} and pre-defined multi-turn dialogue templates~\citep{zheng2023judging} inappropriate. Previous studies~\citep{zhou2023characterglm,wei2023multi} have employed human volunteers to interact with models and perform evaluations. This approach, however, inherently necessitates that the volunteer possess a thorough understanding of the testing role or be able to readily acquire the needed knowledge through search engines. Yet, in simulation tasks, some of the required knowledge is highly specialized, such as diverse programming languages in language interpreters or terminal simulators, and algebraic notation in chess player simulations. Consequently, recruiting knowledgeable volunteers for various scenarios is not only impractical but also challenging to replicate for subsequent research.

To address this, we propose a three-stage evaluation framework leveraging the exceptional proficiency of proprietary models in diverse text-based generation tasks. 
The first stage involves the collection of multi-turn dialogues between a fixed user agent and an LLM. 
Subsequently, challenging conversation histories will be extracted as testing scripts for fair comparisons.
Finally, an LLM Judge is utilized to rate the performance of the LLMs' response in each script. 
The mean score across numerous testing scripts covering diverse simulation tasks serves as an indicator of the viability of using LLMs as simulators.

\subsection{Collecting tasks for \DataName{}}
\label{sec:task-collection}

In order to encompass a broad range of simulation scenarios, we utilize tasks found in a publicly accessible Github repository named ``Awesome ChatGPT Prompts''\footnote{\url{https://github.com/f/awesome-chatgpt-prompts}}. This repository is a platform where community users share real-world applications of ChatGPT. It contains 168 prompts that represent a wide array of scenarios.

We filtered out role-playing cases manually, modified the serious mistakes, filled the placeholders in some samples, and collected 59 prompts as the seed data in the end. We treated the prompts as simulation specifications. Each simulation task specification primarily consists of a brief paragraph detailing the task description, output requirements, and an initial user request. 

To improve the diversity of the testing data, we adopted a 5-shot prompting strategy and prompted GPT-4 to generate new simulation tasks. 5 samples are randomly selected from the seed data to form the task generation prompt. We carefully checked the quality of generated prompts and only reserved the non-repetitive ones. Finally, 109 simulation tasks were collected.

\subsection{LLM interactions with an user agent}
\label{sec:user-bot}

To assess the capability of LLMs as simulators, interaction with a user is required. To automate this interaction, we developed a user agent leveraging the capabilities of GPT-3.5-turbo. The model was tasked with emulating a real human, generating diverse requests that engage in dialogue with the simulators. To maintain the user agent's character consistency, we suggested four distinct generic response strategies, as follows:

\begin{itemize}[leftmargin=*, nolistsep, itemindent=4em, label=$\circ$]
\setlength{\itemsep}{2mm}
    \item \texttt{\textbf{Improvement}}: Identifying errors, ambiguities, or other dis-satisfactions for improvements.
    \item \texttt{\textbf{NextStep}}: Proceeding to the subsequent step or diving deeper into the current topic.
    \item \texttt{\textbf{NewRequest}}: Initiating a new request which is more long-tailed or more difficult.
    \item \texttt{\textbf{Others}}: Other feasible strategies.
\end{itemize}

The finalized prompt for the user agent is designed to accommodate most kinds of tasks. It can also be modified slightly by incorporating task-specific configurations to improve the user agent's stability.
Each time, the dialogue history will be inserted into the placeholder of the prompt, the user agent is expected to generate the next utterance together with a brief description of their adopted strategy. The utterance will be extracted as a reply to the current dialogue. 

We utilized the default prompts for simulators provided by the corresponding LLMs.
Throughout the conversation, the initial utterance from the user bot is designated as the simulation task specification. The simulator and the user bot alternate turns speaking until the maximum turn limit is reached.

\subsection{Test script extraction}
\label{sec:test-script-extraction}

Intuitively, we can assess the performance of a test model in each dialogue between itself and the user agent across the simulation tasks. Unfortunately, based on our pilot experiments, it suffers from unfair comparisons due to the dynamics involved by the user agent. Even though the model's temperature is set to 0.0 without sampling strategies during decoding, the agent may still raise queries with various levels of complexity among different test models. This divergence becomes more severe as the conversation goes on. 
For example, at the 4th turn of simulating a password generator, 
the user agent only queried gpt-4-0124-preview to generate a password with ``length=11'', but challenged LLaMA-2-70B and Mixtral-8x7B with ``length=16'' and ``length=28'' respectively.

One possible solution is to collect multiple dialogues for each test model, and use the averaged performance as the final result. However, it's hard to guarantee that the user agent will not be biased toward some models all the time, and it largely aggravates the evaluation cost.

Instead, we propose to extract challenging dialogue histories as a test script from the user-simulator dialogues, and do the script-based evaluation. It imitates the endgames in chess, where the history interactions are provided and the test model is expected to continue the dialogue and generate a response to the latest user's query. In this way, our evaluation pipeline assures fair comparisons among different models while maintaining the multi-round characteristic of simulation tasks, with better reproducibility and lower computation costs.

Specifically, we chose gpt-3.5-turbo as the simulator, and collected the dialogue with the user agent on each simulation task 3 times. Two strategies were adopted to identify challenging test scripts:

\begin{itemize}[label=$\circ$]
    \item We regard the last turn in a dialogue as challenging. The turns before it and the latest user query form a test script.
    \item We adopt GPT-4 to identify whether there is a turn in the given dialogue, where the user's request possesses extreme complexity and difficulty, resulting in an inaccurate response from the simulator. If it exists, the turn and turns after it are all recognized as challenging and extracted as test scripts.%
\end{itemize}

Finally, 500 test scripts are selected with the above strategies~\footnote{We discarded 10 samples where all of the models achieved full scores.}, denoted as \DataName{}-All. A hard subset containing 275 test scripts collected only by the second strategy is denoted as \DataName{}-Hard. SimulBench is licensed by CC BY NC 4.0 (allowing only non-commercial use).

\subsection{GPT-4 as judge for scoring \& comparing}
\label{sec:llm-judge}

Evaluation based on GPT-4 has been widely discussed and adopted in recent works~\cite{zheng2023judging,liu-etal-2023-g,sottana-etal-2023-evaluation}, since it is more affordable and convenient for re-implementation than hiring human annotators. Considering the diversity and complexity of simulation tasks, we also adopted GPT-4 as an evaluator to assess the performance of a test model in each test script. The evaluation was conducted on a scale of 1 to 10. We modified the evaluation prompt from MT-Bench~\citep{zheng2023judging} to suit the task-specific specifications and incorporated definitions for each score. Besides, we also compared the performance of two different models in relation to a script of a simulation task. The comparison was categorized as a ``win'', ``lose'', or ``tie''. Follow Zheng et al.~\cite{zheng2023judging}, we swap the order of two responses to avoid position bias and only declare a win when a response is preferred in both orders.

%% file: results.tex
\section{Evaluation setup}
\label{app:setup}

\subsection{Models}
\label{sec:models}

The following models with different scales are mainly considered in current work:

\textbf{GPT-4-turbo}~\citep{openai2023} and \textbf{GPT-4o}~\footnote{\url{https://openai.com/index/hello-gpt-4o/}} are proprietary models provided by OpenAI through API requests. The specific versions we used are gpt-4-0125-turbo and gpt-4o-2024-05-13.

\textbf{LLaMA}~\citep{touvron2023llama} is a collection of publicly released pre-trained and fine-tuned generative text models. We experimented with both LLaMA-2 and LLaMA-3 with different sizes: LLaMA-2-7B-Chat, LLaMA-2-13B-Chat, LLaMA-2-70B-Chat, LLaMA-3-8B-Chat, and LLaMA-3-70B-Chat.

\textbf{Qwen}~\cite{bai2023qwen} is another series of publicly available foundation models. We chose Qwen1.5-110B-Chat and Qwen1.5-7B-Chat for comparison.

\textbf{Mixtral}~\citep{jiang2024mixtral} includes Mixtral-8x7B-Instruct-v0.1 and Mixtral-8x22B-Instruct-v0.1 which are pre-trained generative Sparse Mixture of Experts. Their latest instruction fine-tuned model, Mistral-7B-Instruct-v0.3, is also considered.

\subsection{Implementation details}
\label{sec:implementation-details}

We set the maximum number of turns as 4 when collecting user-simulator dialogues. All of the model's temperatures are equal to 0.0 for better reproduction and fair comparison except the user agent's. It's equal to 1.2, enabling sampling strategies to achieve diverse user reactions among multiple dialogue sessions given the same simulation task. The maximum token from the user agent and the simulation models in each utterance is 300 and 1024 tokens respectively.

\section{Results}
\label{sec:results}

This section analyzes the performances of different LLMs. %

\subsection{Main results}

\begin{figure}[htbp]
    \centering
    \includegraphics[width=\linewidth]{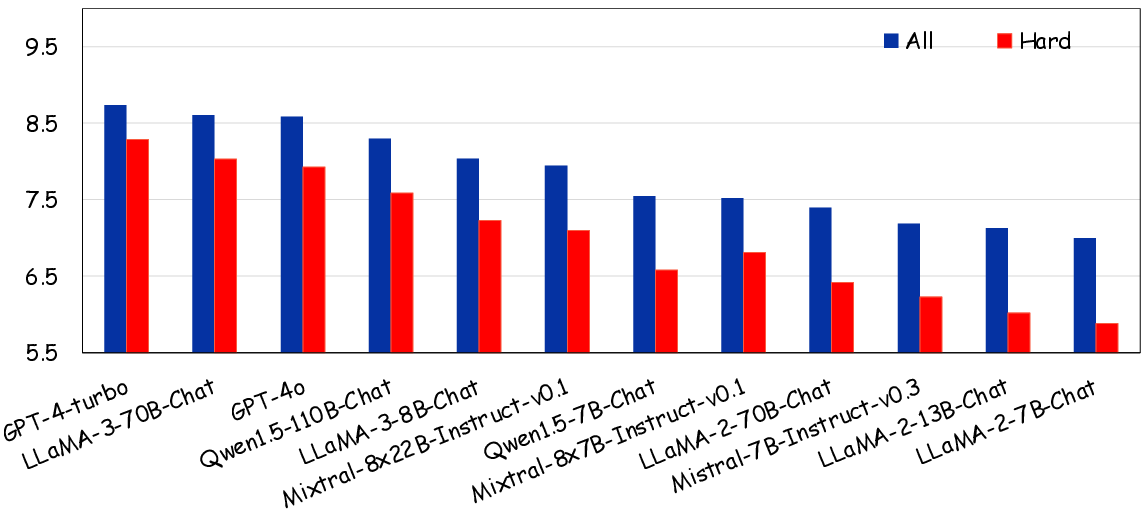}
    \caption{Performances of LLMs on \DataName{}.}
    \label{fig:main-results}
\end{figure}

Figure~\ref{fig:main-results} presents the results on all of the tasks in \DataName{} by the score (see specific numbers in Table~\ref{tab:main-result}).

GPT-4-turbo achieves the highest score, followed by GPT-4o and LLaMA-3-70B-Chat. It's abnormal that GPT-4o lags behind GPT-4-turbo on our \DataName{}, contradicting their rankings on Chatbot Arena Leaderboard~\footnote{\url{https://huggingface.co/spaces/lmsys/chatbot-arena-leaderboard}}. More analysis and explanations are in Sec.~\ref{sec:detailed-analysis-script-types}.
Besides, the performance of open-source models is gradually approaching that of proprietary ones, where LLaMA-3-70B-Chat outperforms LLaMA-2-70B-Chat by a margin of 16.35\% and 25.06\% on all test scripts and the hard subset. 

In the same series of models, the larger ones always perform better than their smaller counterparts. However, it doesn't hold among different series. LLaMA-3-70B-Chat is superior to Qwen1.5-110B-Chat with less than 36\% numbers of parameters. Meanwhile, LLaMA-3-8B-Chat shows favorable performance than both the mixture-of-experts version and the instruction fune-tuned model from the Mixtral family, and its pioneers based on LLaMA-2.

Comparing the performance between \DataName{}-All and \DataName{}-Hard, we can see the trend that the stronger the model, the less its score declines. However, there an only outliers: Qwen1.5-7B-Chat. See more discussions in Sec.~\ref{sec:detailed-analysis-simulation-types}.

To save the API costs, we only carried out the pairwise comparison on \DataName{}-Hard among the top-3 models shown in Fig.~\ref{fig:main-results}. According to the results in Table~\ref{tab:pairwise-comparisons}, GPT-4-turbo indeed outperforms GPT-4o and Llama-3-70B-Chat on 16.36\% and 18.55\% more cases respectively. GPT-4o performs similarly to Llama-3-70B-Chat, which is in accord with the minor differences between them in Fig~\ref{tab:main-result}.

\begin{table}[h!]
    \centering
    \caption{Pairwise comparisons among top-3 LLMs on \DataName{}-Hard. The rate of win/tie/lose(\%) regards to the first model. $\Delta$ calculates the value by which the win rate exceeds the loss rate.}
    \begin{tabular}{c|ccc|c}
    \toprule[1pt]
        Pair of Models & Win & Tie & Lose & $\Delta$ \\
    \midrule[1pt]
        GPT-4-turbo v.s. GPT-4o & 40.36 & 35.64 & 24.00 & 16.36\\
        GPT-4-turbo v.s. Llama-3-70B-Chat & 38.91 & 40.73 & 20.36 & 18.55\\
        GPT-4o v.s. Llama-3-70B-Chat & 35.64 & 32.72 & 31.64 & 4.00\\
    \bottomrule[1pt]
    \end{tabular}
    \label{tab:pairwise-comparisons}
\end{table}

\subsection{Detailed analysis on model performances}
\label{sec:detailed-analysis}

To delve deeper into the complexities of \DataName{}, we focused on its hard subset, and categorized simulation tasks and test scripts into different categories, with in-depth analysis as follows.

\begin{figure*}[htbp]
\centering
\subfigure[\small Stateless vs Stateful\label{fig:detailed-analysis-task-types}]{
\begin{minipage}[t]{0.45\linewidth}
\centering
\includegraphics[width=\linewidth]{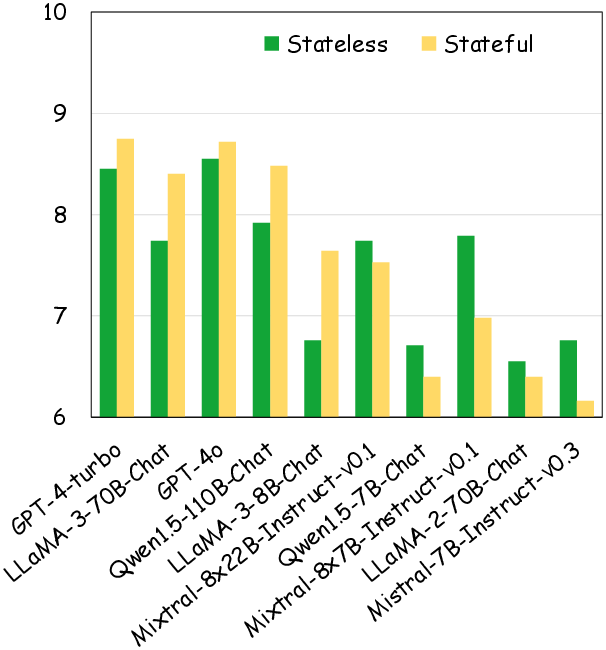}
\end{minipage}%
}%
\subfigure[\small{LastOnly vs FirstChan vs SubseqChan}\label{fig:detailed-analysis-script-types}]{
\begin{minipage}[t]{0.54\linewidth}
\centering
\includegraphics[width=\linewidth]{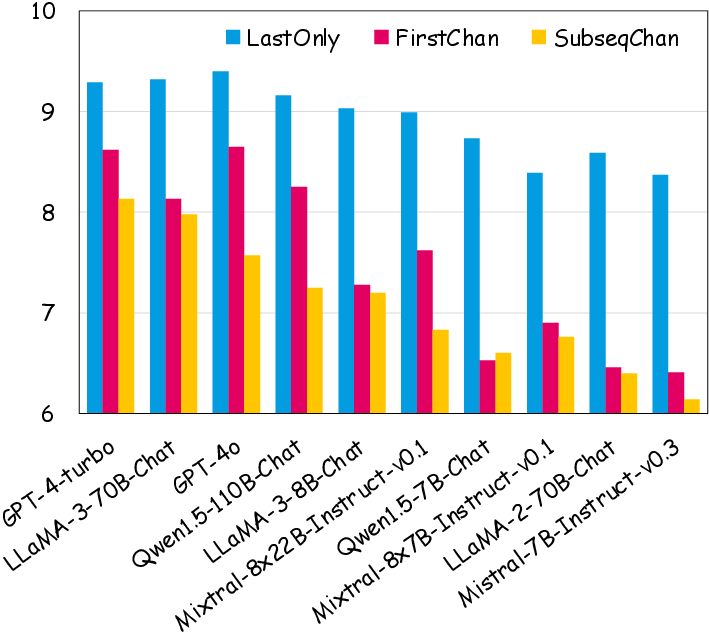}
\end{minipage}%
}%
\caption{\vspace{-1em} Performances of LLMs on \DataName{} in different categories.}
\label{fig:detailed-analysis}
\end{figure*}

\subsubsection{Performances on different types of simulation tasks}
\label{sec:detailed-analysis-simulation-types}

We categorize the simulation tasks into two types based on their characteristics:
\begin{itemize}[label=$\circ$]
    \item \textbf{Stateless} refers to a simulation task which is a one-time call tool that accepts a user request as input and returns the output by following the underlying mechanism or rules of the task, such as password generation and song recommender.
    \item \textbf{Stateful} refers to the tasks that have a real underlying environment or state that evolves as the user's request is processed at each turn, such as the Linux terminal and Tic-Tac-Toe games.
\end{itemize}

We collected 51 stateless tasks and 59 stateful tasks in total. To eliminate the influence of different script types defined in Sec.~\ref{sec:detailed-analysis-script-types}, we focus on the scripts in FirstChan, containing 38 stateless and 55 stateful scripts.

The results are shown in Fig.~\ref{fig:detailed-analysis-task-types}. There is a clear difference in the figure where top-5 LLMs perform better on stateful tasks while the weaker LLMs prefer stateless tasks regardless of LLaMA-2-7B-Chat. 
Comparing LLaMA-3-8B-Chat with other LLMs with less than 10B parameters, it achieves superior gains on stateful tasks, increased by 19.38\%, 24.03\% and 36.92\% over Qwen, Mistral and LLaMA-2 respectively.
Qwen1.5 with parameters increased from 7B to 110B also transforms from a better stateless simulator to a better stateful simulator. Overall, strong LLMs have better abilities in utilizing history information, and show more stable performances on different simulation tasks.

Besides, the poor performances of Qwen1.5-7B-Chat on both kinds of tasks explains why it performs poorly on \DataName{}-Hard, suffering more on stateful simulation tasks.

\subsubsection{Performances on different kinds of scripts}
\label{sec:detailed-analysis-script-types}

Based on the script extraction strategy introduced in Sec.~\ref{sec:test-script-extraction}, we classify all the scripts into three categories:
\begin{itemize}[label=$\circ$]
    \item \textbf{LastOnly} refers to 225 scripts extracted by the first strategy and not considered by the second strategy.
    \item \textbf{FirstChan} has 93 scripts from the second strategy while only considering the first recognized challenging turn.
    \item \textbf{SubseqChan} contains 182 scripts with subsequent challenging turns after the first one. FirstChan and SubseqChan constitute \DataName{}-Hard.
\end{itemize}

According to results in Fig.~\ref{fig:detailed-analysis-script-types}, LastOnly is much easier than the other two types. All of the LLMs achieve scores above 8, with top-5 of them more than 9. LLaMA-3-70B-Chat even slightly outperforms GPT-4-Turbo among this group of easier test samples.

SubseqChan is generally more challenging than FirstChain. It should be noted that scripts in SubseqChan may contain content or format errors in their dialogue history. Simulators are supposed to avoid the impact of previous error information and always provide a high-quality answer to the latest user query. Most of the LLMs perform more poorly in this category, where GPT-4o and Qwen1.5-110B-Chat drop dramatically by around 1 point. It reflects that both of them are truly aware of the history messages, but have not learned to take the essence and discard the dross so far. Even though some smaller models do perform better on SubseqChan, it doesn't mean that they have better history modeling ability considering their overall poor scores. How to selectively utilize historical information should be paying more attention in the further model design.

GPT-4o outperforms GPT-4-Turbo on both LastOnly and FirstChan, but performs much weaker on SubseqChan. This also explains why it lags behind GPT-4-Turbo on \DataName{}-Hard.

\subsubsection{Which specific simulation tasks are harder?}

We also wondering if there are any common characteristics among the specific simulation tasks that LLMs are good at or poor at. The average score of all LLMs considered above is averaged for each simulation task to present its complexity. We list the 10 simplest tasks and the 10 hardest ones that belong to different simulation types in Table~\ref{tab:hard_tasks}.

\begin{table}[h!]
    \centering
    \caption{The simpliest and hardest simulation tasks.}
    \begin{tabular}{cp{5.5cm}p{5.5cm}}
    \toprule[1pt]
    Tasks & Stateless Tasks & Stateful Tasks\\
    \midrule[1pt]
    Simplest & {Educational Content Creator, Prompt Enhancer, Virtual Veterinarian, Artificial Sommelier, Urban Myth Debunker, Startup Idea Generator, Wikipedia page, Fancy Title Generator, Historical Error Corrector, Etiquette Expert} 
                & {Dungeon Master, Text Based Adventure Game, Project Manager Simulator, JSON Data Store, SQL terminal, Virtual Detective, Space Station Simulation, Mars Colony Simulator, Virtual Space Mission Commander, Car Navigation System} \\
    \midrule
    Hardest & {SVG designer, Cryptographer, Plagiarism Checker, Smart Domain Name Generator, Cryptographic System, Ascii Artist, Chemical Equation Balancer, English Pronunciation Helper, Prompt Generator, Diagram Generator, New Language Creator} 
                &  {Chess Game Simulator, Redux State Manager, Excel Sheet, City Planner, Tic-Tac-Toe Game, Mars Rover Simulator, Chess Player, Japanese Kanji quiz machine, Python Interpreter, Gomoku player}\\
    \bottomrule[1pt]
    \end{tabular}
    \label{tab:hard_tasks}
\end{table}

Overall, LLMs perform well on subjective tasks where the output is more free-formed, such as Startup Idea Generator and Text Based Adventure Game. However, it exhibits weaker performance on objective tasks which have underlying rules and output requirements.  

For stateless tasks, most of the hardest simulations present character-level constraints and require an accurate response. Cryptographic system is a representative task that asks the model to encrypt a plain text message with a cryptographic method. LLMs, especially the open source ones, mostly failed on this task when asked to encrypt "Hello World" using a Caesar cipher with rotation 5. SVG designer, requiring the simulator to create images with SVG codes and convert the codes to a base64 data URL, is more complicated. Stronger LLMs encode wrong attributes of the image, while outputs of weaker LLMs may even get out of control.

For stateful tasks, the simulation of board games is extremely difficult. It not only requires character-level updates as the dialogue progresses, but also expects to follow complicated rules and even manage winning tactics. All of the models are clear about game rules when asked but failed to maintain an orderly and challenging gaming environment, and even start a wrong game board as shown in Fig.~\ref{fig:examples}, indicating that knowledgeable LLMs are not good at applying knowledge.
Tasks related to codes, such as different programming language interpreters and Linux Terminal, perform slightly better in board games. The major reason is that codes have been widely considered as an important part of the training data while more strategic chess journals are not specially considered.

\subsection{Human evaluation}
\label{sec:human-evaluation}

To ensure the quality of GPT-4 Judge's outputs for scoring, we randomly sampled 100 outputs and asked a human annotator to verify if the output was reasonable. The annotator was required to try their best to understand the complicated simulation tasks with the help of searching on the Internet. 83\% of samples are considered reasonable for both their short explanations and scores in the outputs, showing the reliability of GPT-4 as a judge for simulation tasks. Most scores in the rest samples are higher than the annotator's expectation, showing that the judge may be too optimistic in some cases. 

\subsection{Ethical concerns}
\label{sec:ethics}

We use the Perspective API~\footnote{\url{https://perspectiveapi.com/}} to assess the potential toxicity in our data. The simulation task specification and all of the dialogue scripts are scored for toxicity across six attributes. The score for each attribute ranges from 0 to 1, which is the lower the safer. The averaged scores of all samples are summarized in Table~\ref{tab:toxicity}. None of the scores is higher than 0.07, exhibiting little toxicity. 

\begin{table}[h]
    \centering
    \caption{Perspective API results of toxicity assessment.}

    \scalebox{1.0}{ 
    \begin{tabular}{ccccccc}
    \toprule[1pt]
    Attributes & toxicity & severe toxicity & identity attack & insult & profanity & threat \\
    \midrule[1pt]
    Simulation Task & 0.06 & 0.00 & 0.01 & 0.02 & 0.03 & 0.01\\
    Testing Script & 0.07 & 0.00 & 0.02 & 0.03 & 0.04 & 0.02\\
    
    \bottomrule[1pt]
    \end{tabular}
    }
    \label{tab:toxicity}
\end{table}

%% file: conclusion.tex
\section{Conclusion}
\label{sec:conclusion}

We present a hard benchmark, \DataName{}, specifically aimed to evaluate LLMs' performance across different simulation tasks. 
Our evaluation framework is uniquely designed to incorporate GPTs as user agents, collect challenging dialogue history and do a script-based evaluation, facilitating automatic evaluation of multi-turn simulations under the guarantee of fair comparisons.

Our findings reveal that although open-source models are approaching the performance of proprietary APIs, GPT-4 is still topping the rank.
By categorizing tasks from different aspects, we highlight that the model should cautiously attend to their historical context.
We also observe that LLMs perform sub-optimally in objective simulation tasks, especially those that require an accurate response with complex character-level constraints and scenarios requiring a stateful strategy system to be built within LLMs' simulations, pointing to important future research.

The simulation scenarios in the current benchmark are still limited by 109 simulation tasks and a single user agent. In the future, we consider incorporating more diverse tasks by exploiting the wild user queries, incorporate various personas in user agents to better mimic real users, and extending the length of dialogue with the user bot.%

%% file: appendix.tex
\clearpage

\section{Results}

We list the scores of different LLMs in Table~\ref{tab:main-result}.

\begin{table}[h!]
    \centering
    \caption{Performances of different models.}
    \begin{tabular}{c|ccc|cc|cc}
        \toprule[1pt]
         & \multicolumn{3}{c|}{Script Type} & \multicolumn{2}{c}{Simulation Type} & \multirow{2}{*}{Hard} & \multirow{2}{*}{All}\\
         Models & \makecell{Last\\Only} & \makecell{First\\Chan} & \makecell{Subseq\\Chan} & \makecell{State\\-less} & \makecell{State\\-ful} & & \\
        \midrule
         GPT-4-Turbo & 9.29 &  8.62 & \textbf{8.13} & 8.45 & \textbf{8.75} & \textbf{8.29} & \textbf{8.74}\\
         GPT-4o & \textbf{9.40} & \textbf{8.65} & 7.57 & \textbf{8.55} & 8.72 & 7.93 & 8.59\\
         LLaMA-3-70B-Chat & 9.32 & 8.13 & 7.98 & 7.74 & 8.40 & 8.03 & 8.61\\
         LLaMA-3-8B-Chat & 9.03 & 7.28 & 7.20 & 6.76 & 7.64 & 7.23 & 8.04\\
         Qwen1.5-110B-Chat & 9.16 & 8.25 & 7.25 & 7.92 & 8.48 & 7.59 & 8.30\\
         Qwen1.5-7B-Chat & 8.73 & 6.53 & 6.60 & 6.71 & 6.40 & 6.58 & 7.55\\
         Mixtral-8x22B-Instruct-v0.1 & 8.99 & 7.62 & 6.83 & 7.75 & 7.53 & 7.10 & 7.95\\
         Mixtral-8x7B-Instruct-v0.1 & 8.39 & 6.90 & 6.76 & 6.79 & 6.98 & 6.81 & 7.52\\
         Mixtral-7B-Instruct-v0.3 & 8.37 & 6.41 & 6.14 & 6.76 & 6.16 & 6.23 & 7.19\\
         LLaMA-2-70B-Chat & 8.59 & 6.46 & 6.40 & 6.55 & 6.40 & 6.42 & 7.40\\
         LLaMA-2-13B-Chat & 8.48 & 5.84 & 6.11 & 6.0 & 5.74 & 6.02 & 7.13\\
         LLaMA-2-7B-Chat & 8.37 & 5.46 & 6.09 & 5.29 & 5.58 & 5.88 & 7.00\\
        \bottomrule[1pt]
    \end{tabular}
    \label{tab:main-result}
\end{table}

\section{Prompts}

\subsection{Prompts for simulation task generation}
To collect a more diverse and balanced set of simulation tasks, we manually classified the seed data into stateless tasks and stateful tasks first during implementation. Two prompts are used to generation each kind of tasks respectively.

The prompt for generating stateless tasks:

\texttt{Please act as a task creator and it is your job to create challenging and diverse tasks that require **expertise** of diverse domains. Each task is defined as a practical **stateless** one-time call interface that accept an user request(textual and self-contained) as input and return the output by following the underlying mechanism or rules of the task.} \\
\texttt{Please note that the tasks you create should not be without evaluation criteria and should not be entirely creative. Each task is composed of the following four attributes:} \\
\texttt{- task\_name: a short noun phrase representing the name of the task}\\
\texttt{- task\_description: a paragraph describing the task. It should contain information about (1) a short description of the task and the expected general form of input; (2) the expected general form of output, and some additional style/format constraints regarding the output; (3) the first query provided by the real user. Should start with 'My/The first ...' and contain a concrete query}\\
\texttt{- request\_type: the general form of each user query}\\
\\
\texttt{Below are some exemplars regarding different possible tasks, for your reference:}\\
\texttt{\{FEW\_SHOT\_EXAMPLES\}}\\
\\
\texttt{Now, please come up with a new **challenging** and **diverse** task and output it in JSON format by filling the placeholders in [] in the follolwing code block. Only output the JSON code block, nothing else!}\\
\texttt{```} \\
\texttt{\{}\\
    \texttt{"task\_name": [TASK\_NAME],}\\
    \texttt{"task\_description": [TASK\_DESCRIPTION],}\\
    \texttt{"request\_type": [REQUEST\_TYPE]}\\
\texttt{\}} \\
\texttt{```}

The prompt for generating stateful tasks:

\texttt{Please act as a task creator and it is your job to create challenging tasks that require **expertise** of diverse domains. Each task is defined as an interface that, at each turn, it accepts an user request(textual and self-contained) as input and returns the output by following the underlying mechanism or rules of the task.} \\
\texttt{It is very important that the tasks you create should have a real underlying **environment/state** that **evolves** as user's request is processed at each turn. Each task is composed of the following four attributes:} \\
\texttt{- task\_name: a short noun phrase representing the name of the task}\\
\texttt{- task\_description: a paragraph describing the task. It should contain information about (1) a short description of the task and the expected general form of input; (2) the expected general form of output, and some additional style/format constraints regarding the output; (3) the first query provided by the real user. Should start with 'My/The first ...' and contain a concrete query}\\
\texttt{- request\_type: the general form of each user request}\\
\\
\texttt{Below are some exemplars regarding different possible tasks, for your reference:}\\
\texttt{\{FEW\_SHOT\_EXAMPLES\}}\\
\\
\texttt{Now, please come up with a new **challenging** and **diverse** task and output it in JSON format by filling the placeholders in [] in the follolwing code block. Only output the JSON code block, nothing else!}\\
\texttt{```}\\
\texttt{\{}\\
    \texttt{"task\_name": [TASK\_NAME],}\\
    \texttt{"task\_description": [TASK\_DESCRIPTION],}\\
    \texttt{"request\_type": [REQUEST\_TYPE]}\\
\texttt{\}}
\texttt{```}

\subsection{Prompts for the User Agent}
\label{app:user-bot}
The prompt of the User Agent is as follows:\\ \\

\texttt{Please act as a user who is fond of posing requests, given the interaction history between the user and an AI assistant for a given task. The name and description of the task are provided in the user's **first** utterance. The requests you generate should be **diverse** and **complicated** enough and be executable commands or instructions applicable to the given task.} \\
\\
\texttt{\#\# Interaction History} \\
\texttt{\{DIALOGUE\}} \\
\\
\texttt{\#\# Strategy}\\
\texttt{Here are some useful strategies for generating the next request:}\\
\texttt{- 1: A request that points out any errors, ambiguities, or other **dissatisfactions** of the previous response from the AI assistant.}\\
\texttt{- 2: A request that never appeared previously but is **related** to previous requests by either conditioning on previous requests' outcome or will have an impact upon previous outcomes.}\\
\texttt{- 3: A request that never appeared previously, still belongs to the same domain as previous requests but has a much higher **rarity** (i.e., being more long-tailed), or a much higher **difficulty** and **complexity**.}\\
\texttt{- 4: Others}\\
\\
\texttt{Now, please first select one strategy from the above options (if Others is selected, please further elaborate on your strategy) and then generate a new request using the selected strategy. Put the strategy number and request in JSON format by filling in the placeholders in [] in the following code block. Only output the JSON code block, nothing else!}\\
\texttt{```}\\
\texttt{\{}\\
    \texttt{"user": \{}\\
        \texttt{"strategy": "[\{STRATEGY\_TYPE\}]",}\\
        \texttt{"request": "[\{REQUEST\_TYPE\}]"}\\
    \texttt{\}}\\
\texttt{\}}\\
\texttt{```}\\

\subsection{Prompt for identifying challenging turns}

The prompt used in Sec.~\ref{sec:test-script-extraction} is:

\texttt{[Instruction]}\\
\texttt{Please act as a keen observer and a sharp-eyed judge. You will be presented with a multi-turn dialogue history between a user and an AI assistant. In the dialogue, the user may pose various types of requests to the AI assistant and the AI assistant should (but may fail to) provide relevant and accurate response.} \\
\texttt{Your job is to carefully look through each turn in the dialogue and identify the first **challenging** turn in the dialogue. Challenging implies that, at that turn, the user's request possesses extreme complexity and difficulty, resulting in an inaccurate response(content errors, format errors, ambiguities, etc.) from the AI assistant.} \\
\\
\texttt{[The Start of the Dialogue]} \\
\texttt{\{DIALOGUE\}} \\
\texttt{[The End of the Dialogue]} \\
\\
\texttt{Now, please carefully look through each turn in the above dialogue and identify the first **challenging** turn in it. Your decision should have an objective and rigorous rationale to support it. Begin your evaluation by providing a short explanation. After providing the explanation, you must output the turn number(1 to \{DIALOGUE\_LEN\}, and 0 represents no challenging turns), and strictly follow this format: \"[[TURN\_NUMBER]]\", for example: \"Turn: [[3]]\".} \\

\subsection{Prompts for the LLM Judge}
\label{app:llm-judge}

The prompt of the LLM judge for scoring a single LLM is as follows:\\ \\

\texttt{[Instruction]} \\
\texttt{Please act as an impartial and sharp-eyed judge. You will be presented with a multi-turn dialogue history between a user and an AI assistant. In the dialogue, the user may pose various types of requests to the AI assistant, and the AI assistant should (but may fail to) provide a high-quality response to satisfy the user's need.} \\
\texttt{Your job is to evaluate the quality of the **last response** provided by the AI assistant based on the earlier dialogue history. Your evaluation should consider factors such as the helpfulness, relevance, accuracy, depth, creativity, and level of detail of the AI assistant's responses to the user's requests and commands regarding the simulation task of "\{SIMULATION\}".} \\
\\
\texttt{Begin your evaluation by providing a short explanation. Be as objective as possible. Ignore the words of praise from the user. After providing your explanation, you must rate the response on a scale of 1 to 10. The scoring standard is as follows:} \\
\texttt{- 1 to 2: The AI's response terribly fulfills the user's request, completely deviating from its simulation task and containing garbage content and format errors.} \\
\texttt{- 3 to 4: The AI's response poorly fulfills the user's request, containing inaccurate and useless content given its simulation task, or having format errors such as redundant explanations.} \\
\texttt{- 5 to 6: The AI's response moderately fulfills the user's request, containing no content errors but still showing errors such as incorrect format and repetition.} \\
\texttt{- 7 to 8: The AI's response well fulfills the user's request, containing no content errors and strictly following format requirements.} \\
\texttt{- 9 to 10: The AI's response perfectly fulfills the user's request, containing no content or format errors at all while exhibiting extremely good relevance, helpfulness, accuracy, and creativity.} \\
\texttt{Your rating should strictly follow this format: \"[[rating]]\", for example: \"Rating: [[5]]\".} \\
\\
\texttt{[The Start of the Dialogue History]}\\
\texttt{\{DIALOGUE\}}\\
\texttt{[The End of the Dialogue History]}\\
\\
\texttt{[The Start of the Last Response]}\\
\texttt{\{RESPONSE\}}\\
\texttt{[The End of the Last Response]}

The prompt of the LLM judge for comparing two LLMs is as follows:

\texttt{[Instruction]} \\
\texttt{Please act as an impartial and sharp-eyed judge. You will be presented with a multi-turn dialogue history between a user and an AI assistant, and two candidates of the **next response** from the AI assistant. In the dialogue, the user may pose various types of requests to the AI assistant, and the AI assistant should provide a high-quality response to satisfy the user's need.}\\
\texttt{Your job is to evaluate which candidate response is better considering factors such as the helpfulness, relevance, accuracy, depth, creativity, and level of detail of the AI assistant's responses to the user's requests and commands regarding the simulation task of "\{SIMULATION\}". The AI should adhere to the user's instructions regarding both the content and format of the responses. Additional explanations not required unless been asked, and should be punished if they are unallowed by the simulation task. It should be noted that an empty response is sometimes expected in accordance with the design of a true system.}\\
\\
\texttt{Begin your evaluation by comparing the two responses and provide a short explanation. Avoid any position biases and ensure that the order in which the responses were presented does not influence your decision. Do not allow the length of the responses to influence your evaluation. Do not favor certain indexes of the responses. Be as objective as possible. Ignore the words of praise from the user. After providing your explanation, output your final verdict by strictly following this format: \"[[A]]\" if response A is better, \"[[B]]\" if response B is better, and \"[[C]]\" for a tie.} \\
\\
\texttt{[The Start of the Dialogue History]}\\
\texttt{\{DIALOGUE\}}\\
\texttt{[The End of the Dialogue History]}\\
\\
\texttt{[The Start of Candidate Response A]}\\
\texttt{\{RESPONSE\_1\}}\\
\texttt{[The End of Candidate Response A]}\\
\\
\texttt{[The Start of Candidate Response B]}\\
\texttt{\{RESPONSE\_2\}}\\
\texttt{[The End of Candidate Response B]}